\documentclass[acmsmall]{acmart}
\usepackage[inline]{enumitem}
\usepackage{multirow}
\usepackage{amsfonts}
\usepackage{float}
\usepackage {graphicx}
  \graphicspath{{./figures/}}
  \DeclareGraphicsExtensions{.pdf,.jpg,.png}
\usepackage{amssymb}
\usepackage{stmaryrd}
\usepackage{adjustbox,lipsum}
\usepackage{tabularx}
\usepackage{array}
\usepackage[font=footnotesize]{subfig}
\usepackage{stfloats}
\fnbelowfloat
\usepackage{url}
\usepackage{verbatim}

\newcommand{\etal}{{\em et al.} }
\newcommand{\eg}{{\em e.g.}, }
\newcommand{\ie}{{\em i.e.}, }
\newcommand{\ri}{{\em i)} }
\newcommand{\rii}{{\em ii)} }
\newcommand{\riii}{{\em iii)} }

\newcommand{\rx}{$\times$ }
\usepackage{booktabs} 
\begin{document}
\title{Face Recognition in Low Quality Images: A Survey}

\author{Pei Li}
\affiliation{\department{Department of Computer Science and Engineering}\institution{University of Notre Dame}}
\author{Patrick J. Flynn}
\affiliation{\department{Department of Computer Science and Engineering}\institution{University of Notre Dame}}

\author{Loreto Prieto}
\author{Domingo Mery}
\affiliation{\department{Department of Computer Sciences}\institution{Pontificia Universidad Cat\'{o}lica de Chile}}

\acmJournal{CSUR}
%${}^2$\hspace{0.5in} Domingo Mery${}^2$\hspace{0.5in} Patrick Flynn${}^1$\\

%${}^1$Department of Computer Science and Engineering\\University of Notre Dame\\
%${}^2$Department of Computer Science\\Pontificia Universidad Catolica de Chile}

\begin{abstract}
Low-quality face recognition (LQFR) has received increasing attention over the past few years. There are numerous potential uses for systems with LQFR capability in real-world environments when high-resolution or high-quality images are difficult or impossible to capture. One of the significant potential application domains for LQFR systems is video surveillance. As the number of surveillance cameras increases (especially in urban environments), the videos that they capture will need to be processed automatically. However, those videos are usually captured with large standoffs, challenging illumination conditions, and diverse angles of view. Faces in these images are generally small in size. Past work on this topic has employed techniques such as super-resolution processing, deblurring, or learning a relationship between different resolution domains. In this paper, we provide a comprehensive review of approaches to low-quality face recognition in the past six years. First, a general problem definition is given, followed by a systematic analysis of the works on this topic by category. Next, we highlight the relevant data sets and summarize their respective experimental results. Finally, we discuss the general limitations and propose priorities for future research.
\end{abstract}

\maketitle

%\IEEEpeerreviewmaketitle

\begin{figure*}[b!]
\begin{center}
\includegraphics[width=\textwidth]{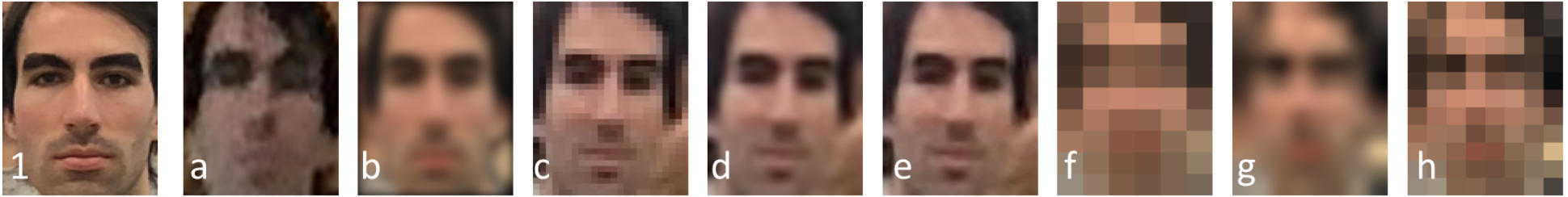}
\caption{Comparison of a high-quality face image (1) with low-quality face images (a--h).}
\label{Fig:LQ_Examples}
\end{center}
\end{figure*}

\section{Introduction}
\label{Sec:Introduction}

Face recognition has been a very active research area in computer vision for decades. Many face image databases, related competitions, and evaluation programs have encouraged innovation, producing more powerful facial recognition technology with promising results. 
In recent years, we have witnessed tremendous improvements in face recognition performance from complex deep neural network architectures trained on millions of face images \cite{parkhi_2015,schroff_2015,sun_2015,wen_2016,liu_2017}.
Although there are face recognition algorithms that can effectively deal with cooperative subjects in controlled environments and with some unconstrained conditions (\eg~\cite{mery_2019}), face recognition in low-quality (LQ) images is far from perfect.
Some examples of LQ face images appear in Fig.~\ref{Fig:LQ_Examples}, illustrating different quality dimensions that can challenge current generation recognition techniques.

LQ face images are produced primarily by four degradation processes applied to high quality inputs:
\begin{itemize}
\item {\em Blurriness} caused by an out-of-focus lens, interlacing, object-camera relative motion, atmospheric turbulence, etc.
\item {\em Low resolution} (LR) caused by a large camera standoff distance and/or a camera sensor with low spatial resolution.
\item  {\em Artifacts} due to low-rate compression settings, motion between fields of interlace-scanned imaging, or other situations.
%can degrade the quality of the image when the compression rate is high.
\item  {\em Acquisition conditions} which add noise to the images (\eg when the illumination level is low).  
\end{itemize}

\noindent Real LQ images can contain all four degradation processes, whereas typically in synthetic LQ images only one process is simulated, \eg LR images are generated by down-sampling and artificial blurriness generated by low-pass filtering of high-quality (HQ) face images. We will cover methods that deal with face recognition under these degradation processes. Thus, the term {\em low-quality} (LQ) face images can include any of the mentioned phenomena, and the recognition will be denoted as {\em low-quality face recognition} (LQFR) in general.

Recent research has begun to address significant challenges to face recognition performance arising from low quality data obtained from surveillance and similar camera installations \cite{li_2018} \cite{li_2017}. The usefulness of automatic face recognition in surveillance data is often motivated by public safety concerns in both public and private sectors. Recognition tasks for such data can include
\begin{itemize}

\item {\em Watch-list identification} -- to determine whether a detected face matches the face of a person on a list of people of interest, or 

\item {\em Re-identification} -- to determine whether a person whose face is captured at one time by one camera, matches a person whose face is captured at a different time and/or from a different camera. 

\end{itemize}
\noindent In these tasks, there are certain scenarios in which the resolution of a face image is high, but the quality is low because of a high amount of blur. Thus, it is possible to have a HR image that is LQ (see, for example, in Fig. \ref{Fig:LQ_Examples}b a LQ face image with HR, 110 \rx 90 pixels).

We denote  those images that have not been degraded by any of the mentioned degradation processes as high-quality (HQ) images. The gallery images of known persons that are used in the above-mentioned tasks can often be HQ frontal images.  However, the probe images are usually LQ: they have been taken from surveillance cameras delivering small faces, potentially in blurry images of poorly lit surroundings from a viewpoint elevated well above the head. This situation is illustrated in Fig. \ref{Fig:LQ_Examples}.
%, in which the matching HQ$\leftrightarrow$LQ, see for example faces `1' and `f', is established.
Most classifiers designed to work well with high-quality face images in both gallery and probe set cannot properly handle a LQ probe due to the quality mismatch.

This paper captures the research in LQFR since 2012. The reader is referred to Wang \etal~\cite{wang_2014}, that provided a comprehensive summary of works in this area prior to 2012, in which classical methods (developed before the advent of deep learning) are discussed. In our work, we compare the most recent approaches as well as new data sets used for evaluation. 

Research works are discussed in the following parts. In Section~\ref{Sec:LQFR}, a description of face recognition in low-quality images is given mentioning the problems, solutions and human performance. In Sections~\ref{Sec:SR}, \ref{Sec:LowResolutionRobustFeature}, and \ref{Sec:UnifiedSpace}, \ref{Sec:Deblurring}, existing computer vision methods in this field are presented in four categories: (i) super-resolution based methods, (ii) methods employing low-resolution robust features, (iii) methods learning a unified representation space, (iv) remedies for blurriness. Within each section, the methods are reviewed either by publication or, where possible, grouped by general approach. We also summarize the state-of-the-art deep learning-based methods for insight on their similarities and comparison with the traditional methods. In Section~\ref{Sec:data setEvaluations}, data sets and evaluation protocols that are used in this research area are outlined. Finally, in Section~\ref{Sec:Conclusions}, concluding remarks, trends, and future research are addressed.

\section{Face recognition in low-quality face images}
\label{Sec:LQFR}
It is clear that face recognition, performed by machines or even by humans, is far from perfect when tackling LQ face images. In this section, we  give a general overview of the face recognition problem in low-quality images. We address the differences between LQ and HQ face recognition (Section \ref{Sec:LQHQ}), present a definition of LR face images (Section \ref{Sec:LR_definition}), summarize the existing works that study human performance on the LQFR task (Section \ref{Sec:HumanVision}), and highlight the potential challenges noted in related LQFR research (Section \ref{Sec:challenge}).

\begin{figure}[t!]
\centering
\includegraphics[width=1.0\columnwidth]{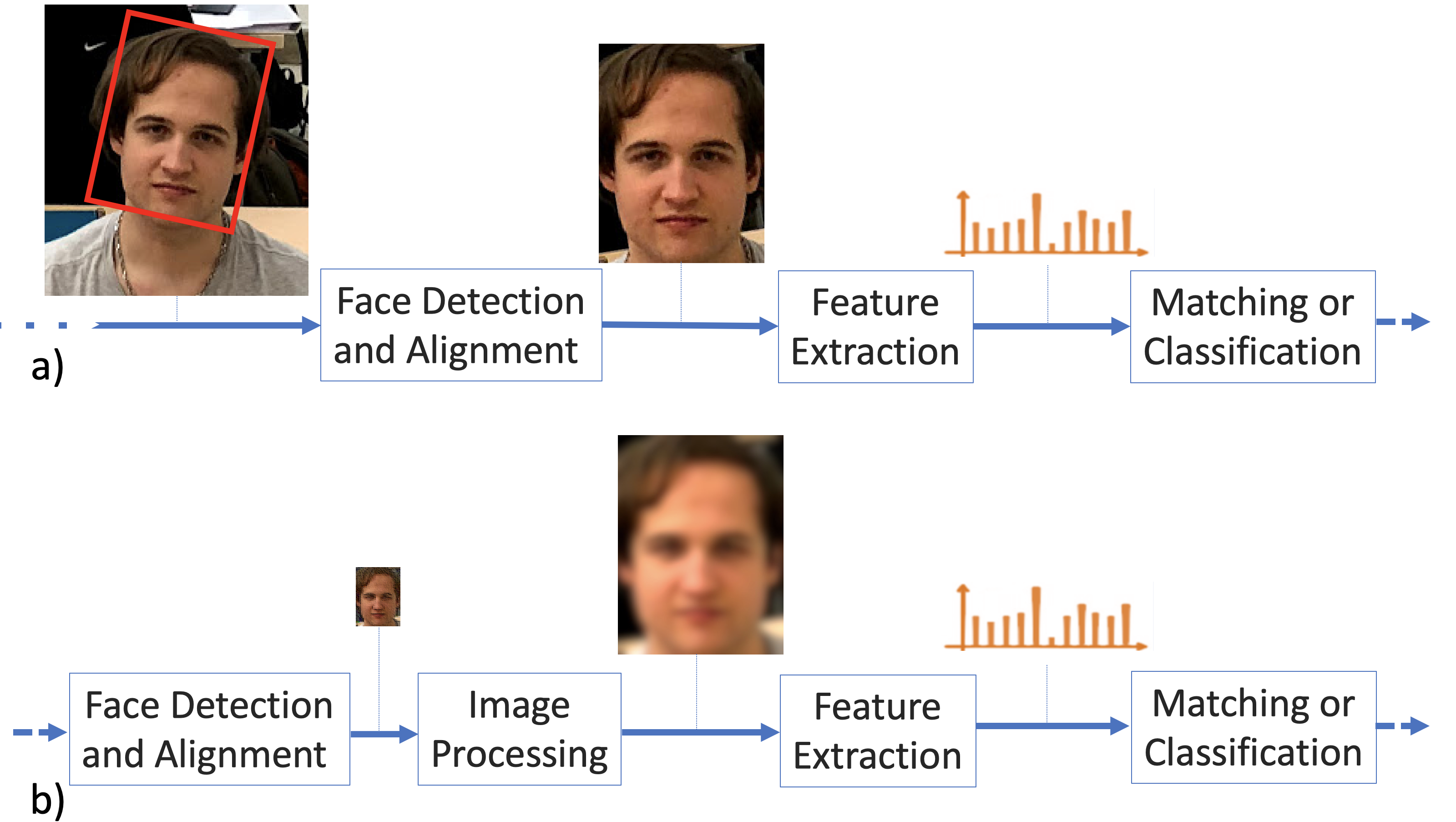} 
\caption{General schema for face recognition in a) HQ and b) LQ. Typically in LQFR, a step before feature extraction is included; In general, this step is called "image processing". For example, image processing could include a super-resolution approach in order to improve the quality of the input image.}
\label{Fig:Schema_HQ_vs_LQ}
\end{figure}

\subsection{LQ vs. HQ face recognition}
\label{Sec:LQHQ}
Face recognition has been carefully summarized in several survey papers that address HQ images exclusively or primarily \cite{wagner_2012,zhao_2003,bowyer_2006,ding_2016,jafri_2009,galbally_2014}. Existing approaches typically follow the schema of Fig. \ref{Fig:Schema_HQ_vs_LQ}a: face detection, face alignment, feature extraction and matching or classification. Extracted features must be discriminative enough, \ie features from two different face images from same (or different) subjects must be very similar (or dissimilar). Face recognition performance has improved significantly in recent years, to a level comparable to that of humans \cite{parkhi_2015,schroff_2015,sun_2015} in high-quality face images. 

Face recognition systems are now deployed on a large scale and their use is expanding to include mobile device security and the retail industry as well as  public security. Examples of the latter application domain include long-distance surveillance and person re-identification  \cite{li_2017,li_2018}, in which severely blurred and very low-resolution images (\eg face images of 10 \rx 10 pixels) yield considerable deterioration in recognition performance. Compared to face recognition in a controlled environment with stable pose and illumination, face recognition under less constrained scenarios performs relatively poorly due to the low quality of the face images captured. The cameras used for surveillance usually have limited resolution and are far from the subjects. Faces in images captured by surveillance devices usually lack high frequency or detailed discriminative features and can be blurry due to defocus and subject and/or camera motion.

\begin{figure}[t!]
\centering
\includegraphics[width=\columnwidth]{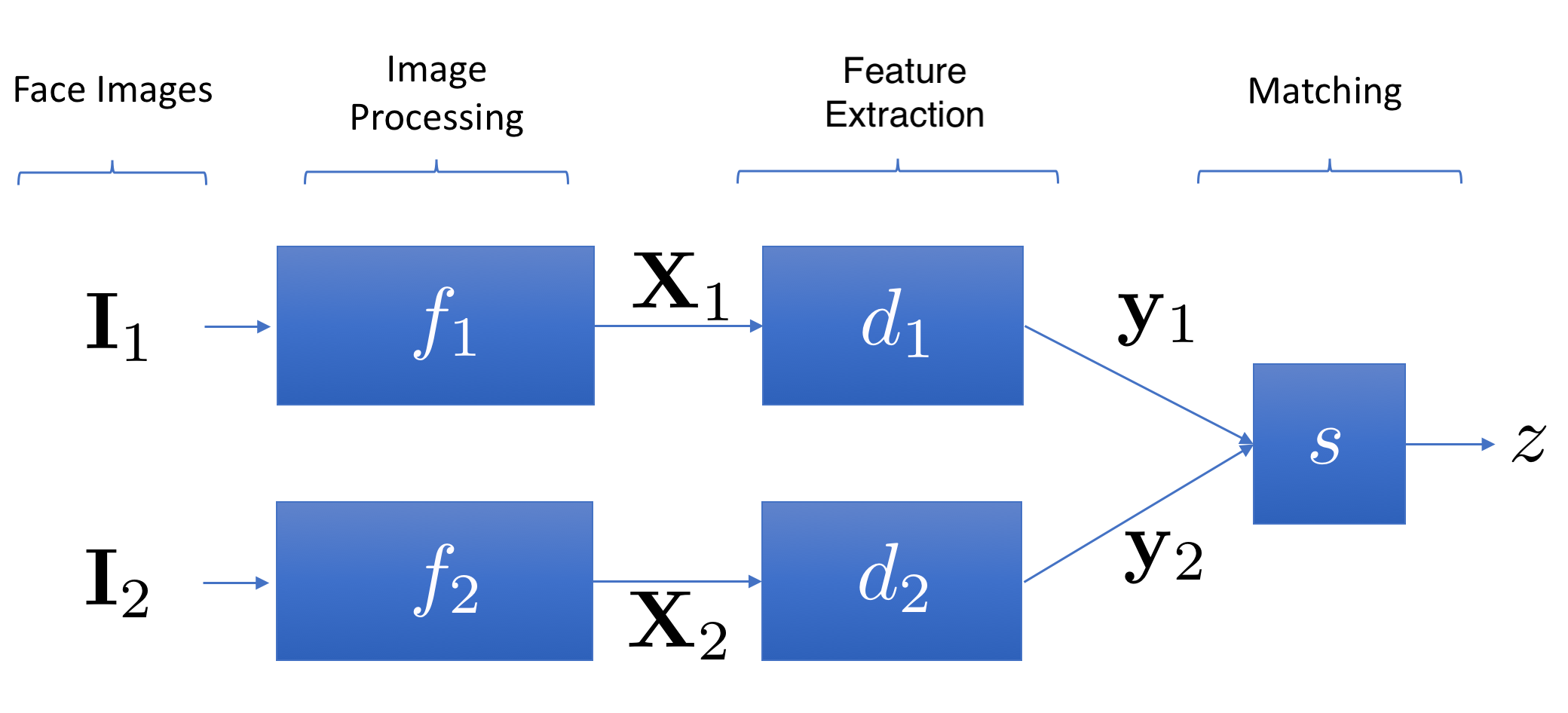} 
\caption{General schema for LQFR for matching two images.}
\label{Fig:Schema_LQFR_Matching}
\end{figure}

Unlike face recognition in HQ images, face recognition in LQ images is a very challenging task in all of the pipeline steps (see Fig. \ref{Fig:Schema_HQ_vs_LQ}b). Face detection applied to LQ images acquired in surveillance is the first challenge. State-of-the-art methods \cite{najibi_2017,zhang_2016,zhang2017s} and large face detection challenges like FDDB \cite{jain_2010} and WIDER FACE \cite{yang2016wider}, achieve excellent results with face images typically larger than 20 \rx 20 pixels. Another challenging stage is face alignment \cite{wagner_2012,liu_2017b,peng_2017}: most face landmark detectors are trained on HQ face images with distinct landmarks, and can be expected to fail when applied on LQ face images. Misalignments degrade to performance due to the small size of the face region. The general schema for matching in LQFR is illustrated in Fig. \ref{Fig:Schema_LQFR_Matching}, where two face images (${\bf I}_1$ and ${\bf I}_2$) are to be matched. The input images can have different qualities, for instance ${\bf I}_1$ is LQ and and ${\bf I}_2$ is HQ. The general `image processing' stage (functions $f_1$ and $f_2$), before the `feature extraction' step, is usually included to improve the quality, \eg it can change the resolution of the input images. There are many approaches that attempt to improve the quality of a LQ face image. In the case of image restoration methods, super-resolution techniques have been developed. If the quality of input image is high enough, no image processing is required. In general, the new representations ${\bf X}_1$ and ${\bf X}_2$ can be in the image space or in another one. The idea is to extract features, using functions $d_1$ and $d_2$ to obtain descriptors ${\bf y}_1$ and ${\bf y}_2$, that can be matched by a similarity function $s({\bf y}_1,{\bf y}_2)$. Input images are matched (or not matched) if the similarity is high (or low). 

In order to tackle the face recognition problem in LQ images, `real low-resolution images' (acquired by real cameras as mentioned above) and  'synthetic low-resolution images' (obtained from high-quality inputs by subsampling, blur application, and other operations intended to degrade quality in a parameterized way) are used for training and testing purposes. Real images are not only lower in resolution but also have a higher noise level and deviate in other ways from the high-resolution gallery images. 

\subsection{Resolution as an Element of Quality}
\label{Sec:LR_definition}
Although there is no broadly accepted single criterion for labeling a face in an image as {\em low-resolution}, many works have recognized that face images with a tight bounding box smaller than 32 \rx 32 pixels begin to present significant accuracy challenges to face recognition systems in both human and computer vision (see for example \cite{boom_2006}). Other works, however, mentions that the minimal resolution that allows identification is 16 \rx 16 (see for example \cite{harmon_1973}). Occasionally, the inter-ocular distance (as measured by pupil center distance or, incompatibly, distance between inner or outer eye corners) is used instead. Wang \etal \cite{wang_2014} defined two concepts:

\begin{itemize}

\item {\em The best resolution} -- ``on which the optimal performance can be obtained with the perfect trade-off between recognition accuracy and performing speed'' and 

\item {\em The minimal resolution} -- ``above  which  the  performance  remains  steady  with the resolution decreasing from the best resolution, but below which the performance deteriorates rapidly''.
\end{itemize}
\noindent They concluded that the best resolution is not always the highest resolution that are able to obtain and the minimal resolution depends on different methods and databases.

Xu \etal \cite{xu_2014} investigated three important factors that influence face recognition performance: \ri type of cameras (high definition cameras {\em vs.} surveillance cameras), \rii the standoff between the object and camera, and \riii how the LR face images are collected (camera native resolution or by downsampling images of larger faces). They compared FR performance with faces of different qualities obtained from high definition cameras and surveillance at different standoffs. They discovered that the down-sampled face images are not good representations of captured LR images for face recognition: the performance continued to degrade when the resolution decreased, which occurred in both of the standoff scenarios.

Some researchers have noted the dependence of a `minimum useful resolution' on the face recognition technology in use. Marchiniak \etal \cite{marciniak_2015} presented the minimum requirements for the resolution of facial images by comparing various face detection techniques. In addition, they also provided an analysis of the influence of resolution reduction on the FAR/FRR of the recognition.

Extended works have been carried out in recent years to determine principled techniques for assessing the resolution component of image quality. 
Peng \etal \cite{peng_2017} demonstrated the difference in recognition performance between real low-resolution images and `synthetic' low-resolution images down-sampled from high-resolution images. Phillips \etal \cite{phillips_2013} introduced a greedy pruned ordering (GPO) as an approximation to an image quality oracle which provides an estimated upper bound for quality measures. They compared this standard against 12 commonly proposed face image quality
measures. Kim \etal \cite{kim_2014} proposed a new automated face quality assessment (FQA) framework that cooperated with a practical FR system employing candidate filtering. It shows that the cascaded FQA can successfully discard face images that could negatively affect FR. They demonstrated that the recognition rate on the FRGC 2.0 data set increased from 90\%  to 95\% when unqualified faces are rejected by their methods.

\subsection{Human vision in LQFR}
\label{Sec:HumanVision}
Human visual object recognition is typically rapid and seemingly effortless \cite{tanaka_2016}. It is treated as the benchmark of choice when comparing to manually designed algorithms \cite{phillips_2018}. Face recognition can be an easy task for humans in simple cases \cite{harmon_1973}; however, there is clear evidence that in more challenging scenarios it is a difficult and error-prone task \cite{zeinstra_2017,sinha_2006}. Although the state-of-the-art machine learning models such as deep convolutional networks have achieved human-level classification performance on some tasks, they do not currently seem  capable of performing well on the recognition problem in LQ images. 

Several human studies conducted on LQFR have been conducted. In the 1990s, Burton \etal \cite{burton_1999} examined the ability of subjects to identify target people captured by a commercially available video security device. By obscuring face, gait, and body, they conclude that subjects were using information from the face to identify people in these videos. There was a smaller reduction in accuracy when a person's gait or body was concealed compared to when face was concealed. This indicates that face (if available) is a more important clue for identifying people's identity in LQ video. Best-Rowden \etal \cite{bestrowden_2014} reported unconstrained face recognition performance by computer and human beings, and proposed an improved solution combining humans and a commercial face matcher. Robertson \etal \cite{robertson_2016} evaluated the performance of four working human `super-recognizers' within a police force and consistently find that this group of persons performed at well above normal levels on tests of unfamiliar and familiar face matching, even with size-reduced images ($30 \times 45$ pixels). The human super-recognizer achieved 93\% accuracy on a familiar face matching task containing 30 famous celebrities in 60 different matching trials. 

In order to assess human face recognition accuracy on low-resolution images, six experiments have been conducted with different degrees of familiarity using 30 persons to identify (celebrities that can be familiar or unfamiliar to the viewers) \cite{noyes_2016}. Unsurprisingly, the higher the resolution and the familiarity, the better the accuracy was. In addition, low-resolution images can be better recognized if they are blurred using a low-pass filter. Using a {\em wisdom-of-crowds} strategy, the accuracy is increased using a group average response. In addition, Noyes \etal \cite{noyes_2017} discovered that changing camera-to-subject offset (0.32m versus 2.70m) impaired perceptual matching of unfamiliar faces, even though the images were presented at the same size. However, the matching performance on familiar faces was accurate across conditions, indicating that perceptual constancy compensates for distance-related changes in optical face shape. 

Historical evidence showed that human face recognition is very robust against some challenges (\eg illumination, pose, etc.) and very poor against some others (\eg low-resolution, blurring, time lapse between images, etc.). To establish the frontiers of LQFR in both human and computer vision is an open question, however, it seems to be possible that they can mutually benefit from one another.

\subsection{Challenges in LQFR}
\label{Sec:challenge}

 In summary, LQFR has following challenging attributes: 
https://www.overleaf.com/project/59b6d65c58adb24292f1e5d0
\begin{itemize}

\item {\em Small face region and lack of details:} Some of the faces captured by normal surveillance cameras are low-resolution. Some video frames are captured by high-resolution cameras but from large standoffs. The faces in such images lack local details which are essential for face recognition. 

\item {\em Blurriness:} Some faces in LQ images are contaminated by  blurriness caused by either poor focus or movement of the subject. 

\item {\em Non-aligned:} Face landmark detection is highly challenging on low-resolution faces which results in automatic face alignment failure or inaccuracy.

\item {\em Variety in dimension:} As a common surveillance task, the target subject could walk in a completely unconstrained way, getting closer or further to the camera with a random pose relative to the camera's optical axis. This leads to various sizes of faces detected. When there is a large difference in the dimension, face matching performance might be influenced. 

%% Not sure what this item is trying to say.
\item {\em  Scarce data sets:} The number of data sets designed to provide low-quality face images is very limited.

\end{itemize}

%%%%%%%%%%%%%%%%%%%%%%%%%%%%%%%%%%%%%%%%%%%%%%%%

%\section{Methods}
%\label{Sec:Methods}
%We illustrate several notable published works for LQFR in five general classes: (i) super-resolution based methods, (ii) methods employing low-resolution robust features, (iii) methods learning a unified representation space, (iv) remedies for blurriness,  and (v) deep learning based method. They will be reviewed in Sections \ref{Sec:SR}, \ref{Sec:LowResolutionRobustFeature}, \ref{Sec:UnifiedSpace}, \ref{Sec:Deblurring} respectively.%Deep learning, as an effective way to learning good representations over various tasks, also achieves impressive results on LQFR problem over classical modeling approaches. We list all the methods employing deep learning separately, since they share similarities in designing and optimization. 

\section{Super-resolution}
\label{Sec:SR}
\subsection{Background}
The accuracy of traditional face recognition systems is degraded significantly when presented with LQ faces. The most intuitive way to enhance the quality of these face images as an input of the face recognition system or a human face recognizer is super-resolution (so that more informative details can be perceived). State-of-the-art face super-resolution (SR) methods have been proposed over the past few years, which boost both the performance on visual assessment and recognition. % AMY merge this into SR section

In this section, we  introduce two categories of approaches to super-resolution processing. One category focuses on reconstructing visually high-quality images, the other employs super-resolution methods for face recognition explicitly.  There is also another way to categorize these super-resolution methods, as either  reconstruction based or learning based (as noted by Wang \etal state in \cite{wang_2014}) based on whether the process incorporated any specific prior information during the super-resolving process. Learning based approaches generally produce better results and have dominated the most recent work.  
Xu \etal \cite{xu2013face} investigated how much face-SR can improve face recognition，reaching the conclusion that when the resolution of the input LR faces is larger than 32 $\times$ 32 pixels, the super-resolved HR face images can be better recognized than the LR face images. However, when the input faces have very low dimension (e.g., 8 $\times$ 8 pixels), some of the face-SR approaches do not work properly and therefore little improvement in performance can be expected.
 
\subsection{Methods}
\subsubsection{Visual quality driven face super-resolution}
\label{Sec:VisualQuality_DFSR}
Resolving LR faces to HR faces introduces more details on the pixel level which is potentially good for boosting the recognition rate; however, it might introduce artifacts as well. One way of assessing the quality of the resolved LR faces is by using the human visual system. The most common quantitative metrics for measuring the reconstruction performance are peak-signal-to-noise ratio (PSNR) and structural similarity index measure (SSIM). For recognition purpose, using the super-resolved face images for matching usually achieves better performance compared to directly matching the LR face probes with HR face gallery images.

Sparse representation based methods employing patch decompositions of the input images are used widely for super-resolution LR face images by assuming that
each (low-resolution, high-resolution) patch pair is sparsely represented over the transformation space where the resulting representations correspond.
Earlier work on image statistics discovered that image patches can be well-represented as a sparse linear combinations of elements from an appropriately chosen over-complete dictionary. By jointly training two dictionaries for the LR and HR image patches with local sparse modeling, more compact sparse representations were obtained. 

\begin{figure*}[ht]
\centering
\includegraphics[width=13cm]{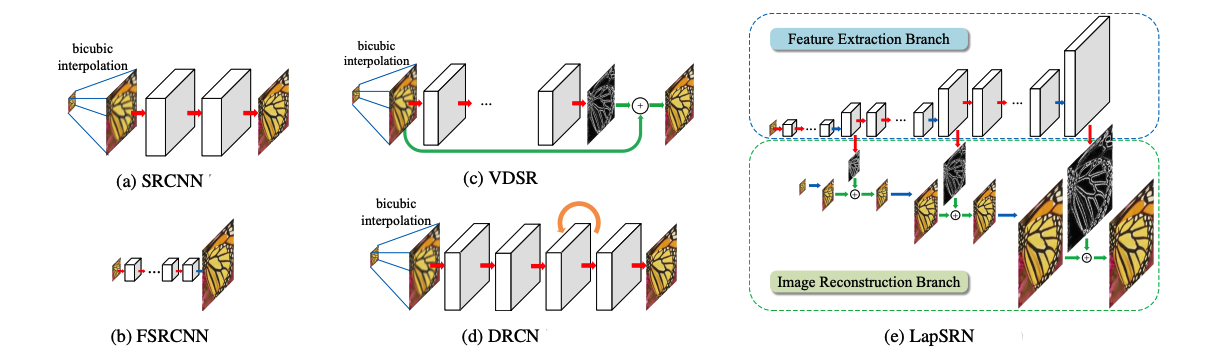}
\caption{General deep learning-based SR architectures. (Figure taken from \cite{lai2018fast})}
\label{Fig:dsrarchs}
\end{figure*}

The features are  robust to resolution change and further enhance the edges and textures in the recovered image. Wang \etal \cite{2015:Wang} proposed an end-to-end CNN based sparse modeling method that achieved better performance by employing a cascade of super-resolution convolutional nets (CSCN) trained for small scaling factors. Jiang~\etal~\cite{jiang2014noise,jiang2017noise} was able to alleviate noise effects during the sparse modeling process by using a locality-based smoothing constraint. They proposed an improved patch-based method ~\cite{jiang2017context} which leverages contextual information to develop a more robust and efficient context-patch face hallucination algorithm. Farrugia~\etal \cite{2017:Farrugia} estimated the HR patch from globally optimal LR patches. A sparse coding scheme with multivariate ridge regression was used in the local patch projection model in which the LR patches are hallucinated and stitched together to form the HR patches.
Jiang~\etal~\cite{2014:Jiang,jiang2016facial} employed a multi-layer locality-constrained iterative neighbor embedding technique to super-resolve the face images from coarse-to-fine scale as well as preserving the geometry of the original HR space. Similarly, Lu~\etal~\cite{lu2017robust} used a locality-constrained low-rank representation scheme to handle the SR problem.  
Xu~\etal~\cite{2015:Juefei} modeled the super-resolution task as a missing pixel problem. This work employed a solo dictionary learning scheme to structurally fill pixels into LR face images to recover its HR face counterpart. Jiang~\etal~\cite{jiang2017srlsp} also formed the super-resolution task as interpolating pixels in LR face images using smooth regression on local patches. The relationship between LR pixels and missing HR pixels was then learned for the super-resolution task.
\paragraph*{Deep learning}
Some state-of-the-art approaches employed deep learning or a GAN model to handle the super-resolution task.
There are several influential state-of-the-art deep architectures for general image super-resolution, as shown in~\ref{Fig:dsrarchs}.
%Most of the deep learning based architectures designing their models drew the experience from them. 
Huang~\etal~\cite{huang2017wavelet} was inspired by the traditional wavelet that can depict the contextual and textural information of an image at different levels, and designed a deep architecture containing three parts (feature map extraction, wavelet transformation, and coefficient prediction), and reconstructed the HR image based on predicted wavelet coefficient. Three losses (wavelet prediction loss, texture loss, and full image loss) were combined in the training process.

\begin{figure*}[ht]
\centering
\includegraphics[width=13cm]{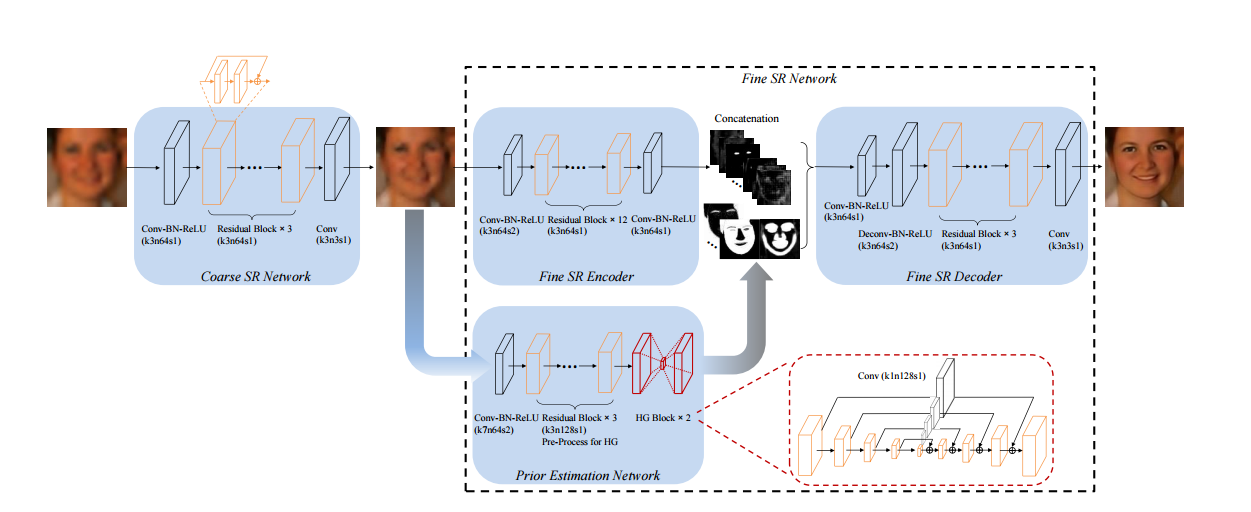}
\caption{FSRNet architecture. (Figure taken from~\cite{chen2017fsrnet})}
\label{Fig:FSRNet}
\end{figure*}

\begin{figure*}[ht]
\centering
\includegraphics[width=13cm]{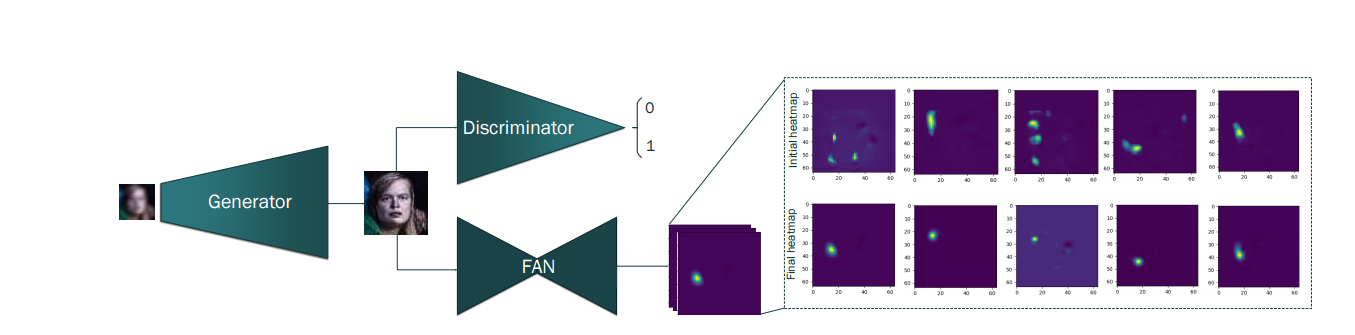}
\caption{Super-FAN network architecture. (Figure taken from~\cite{bulat2017super})}
\label{Fig:SuperFAN}
\end{figure*}

\begin{figure*}[ht]
\centering
\includegraphics[width=13cm]{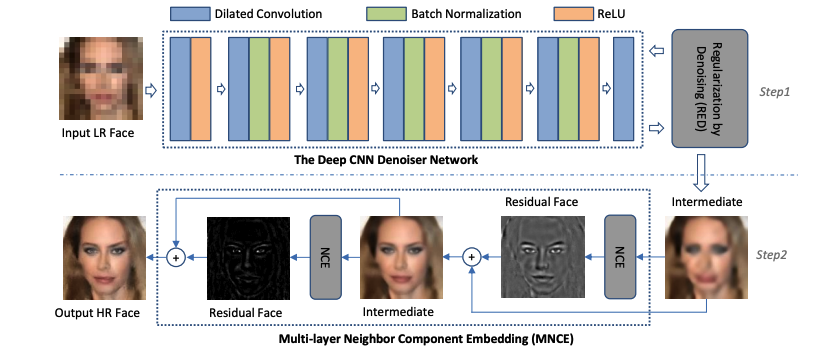}
\caption{Super-FAN network architecture. (Figure taken from~\cite{jiang2018deep})}
\label{Fig:Dnoise}
\end{figure*}

Facial landmark detection for face alignment and super-resolution is a chicken-and-egg problem and many approaches were not able to address both of the aspects. Recently, some deep learning-based methods try to incorporate facial priors to improve super-resolution performance.
Chen~\etal~\cite{chen2017fsrnet} proposed a method (depicted in Figure~\ref{Fig:FSRNet}) with an architecture that contains five components: a coarse SR network, a fine SR encoder, a prior estimation network, a fine SR decoder, and a GAN for face realization. The face shape information is better preserved compared to the texture, and is more likely to facilitate the super-resolution task. They integrate an hourglass structure to estimate facial landmark heatmaps and parse the maps as facial priors which are concatenated with intermediate features for decoding HR face images. Similarly, Bulat~\etal's method \cite{bulat2017super} (shown in Figure \ref{Fig:SuperFAN}) proposed a GAN-based architecture which also included a facial prior that is achieved by incorporating a facial alignment subnetwork. They introduced a heatmap loss to reduce the difference between feature heatmaps in the original HR face images and those in the LR face images. In this way, they achieved rough facial alignment utilizing the information from the feature map. Both of the two works mentioned above tried to incorporate facial structure as prior in a weak supervised manner to improve the super-resolution and in turn to improve the prior information while in the training process.
Similar to ~\cite{chen2017fsrnet}, Jiang \etal~\cite{jiang2018deep} employed the CNN-based denoised prior within the super-resolution optimization model with the aid of image-adaptive Laplacian regularization, as shown in Figure~\ref{Fig:Dnoise}. They further develop a high frequency detail compensation method by dividing the face image to facial components and performing face hallucination in a multi-layer neighbor embedding approach. 
In 2017, Yu~\etal~\cite{yu2017face} presented an end-to-end transformative discriminative neural network (TDN) employing the spatial transformation layers to devise super-resolving for unaligned and very small face images. In the following year, they developed an attribute-embedded upsampling network~\cite{yu2018super} utilizing supplementing residual images or feature maps that came from the difference of the HR and LR images. With additional facial attribute information, they significantly reduce the ambiguity in face super-resolution. Chen \etal~\cite{chen2017face} showed a comparison of the performance of three versions of GANs in the context of face superresolution: the original GAN, WGAN, and the improved WGAN. In their experiments, they evaluated the stability of training and the quality of generated images. 

\subsubsection{Face super-resolution for recognition}
\label{Sec:FaceSR_forRecognition}
Since most of the face super-resolution methods generate a visual result (a super-resolved HR face output), the performance can be measured using visually quantitative metrics, as described in Section~\ref{Sec:VisualQuality_DFSR}. There are some other works that focus on using super-resolution to enhance LQFR performance. These methods all reported comparable or superior LQFR performance on standard public data sets. The following subsections introduce these approaches in two groups. One group of methods takes the visually enhanced super-resolution output as the first step and executes a subsequent recognition process. The other group incorporates super-resolution into the recognition pipeline by trying to regularize the super-resolution process to generate robust intermediate results or features for recognition. Compared with the system doing super-resolution and recognition sequentially, this second group of approaches emphasized the incorporation of inter-class information during the learning process to obtain more discriminative features that are directly related to the classification task. Some of these methods were able to optimize the parameters for feature extraction part as well as the classifier together during training. 

\paragraph{Super-resolution-aided LQFR}
\label{Sec:SR_aided_LQFR}

Xu~\etal~\cite{xu_2014} combined the work of Yang~\etal~\cite{2010:Yang} and Wang~\etal~\cite{wang2005hallucinating} and employed dictionary learning for patch-based sparse representation and fusion from super-resolved face image sequences.
The work of Uiboupin~\etal~\cite{uiboupin2016facial} proposed the learning of sparse representations using dictionary learning. However, to get more realistic output, they used two different dictionaries -- one trained with natural and facial images, the other with facial images only. They employ a Hidden Markov Model for feature extraction and recognition.
Aouada~\etal's work ~\cite{2014:Aouada}, proceeding from the work of Beretti~\etal~\cite{berretti2012superfaces}, addressed the limitation of using a depth camera for face recognition. It stated that the 3D scans can be processed to extract a higher resolution 3D face model. By performing a deblurring phase in the proposed 3D super-resolution method, the approach boosted the recognition performance on the Superfaces data set~\cite{berretti2012superfaces} from 50\% to 80\%.  

Canonical correlation analysis (CCA) has been widely adopted for manifold learning and showed its success when applied to recognition-oriented face super-resolution. However, 1D CCA was not designed specifically for image data. To fit the image data into a 1D CCA formulation, the image has to be first converted into a 1D vector which might obscure appearance details in the image. To overcome this, An~\etal~\cite{2014:An} proposed a 2D CCA method that took two sets of images and explores their relations directly without vectorizing each image. It performed super-resolution in two steps: face reconstruction and detail compensation for further refinement with more details. It achieved more than 30\% of performance improvement on a hybrid data set constructed from the CAS-PEAL-R1 and CUHK data sets. In the same year, Pong~\etal~\cite{2014:Pong} proposed a method to enhance the features by extracting and combining them at different resolutions. They employed cascaded generalized canonical correlation analysis (GCCA) to fuse Gabor features from three levels of resolution of the same image to form a single feature vector for face recognition.
Jia~\etal~\cite{2016:Jia} also employed CCA to establish the coherent subspaces for HR and LR PCA features. The super-resolution process happened in the feature space in which the LR PCA features were hallucinated using adaptive pixel-wise kernel partial least squares (P-KPLS) predictor fused with LBP features to form the final feature representation.
Satiro~\etal~\cite{2015:Satiro} used motion estimation on interpolated LR images, and employed the resulting parameters in a non-local mean-based SR algorithm to produce a higher quality image. An alpha-blending approach is then applied to fuse the super-resolved image with the interpolated reference image to form the final outputs. 
\paragraph*{Deep learning}
Wang et al.~\cite{2015:Wang} proposed a feed-forward neural network whose layers strictly correspond to each step in the processing flow of sparse coding-based image SR. All the components of sparse coding can be trained jointly through back-propagation. They initialized the parameters correspondingly to the understanding of different components of the classical sparse coding method. 
Wang~\etal~\cite{2016:Wang} proposed a semi-coupled deep regression model for pretraining and weight initialization. They then fine-tuned their pre-trained model with different data sets including a low-resolution face data set which achieved outstanding performance. Similar to Juefei-Xu~\etal~\cite{2015:Juefei}, Jiang~\etal~\cite{jiang2017srlsp} learned the relationship between LR pixels and missing HR pixels of one position patch using smooth regression with a local structure prior, assuming that face image patches at the same position share similar local structures. They reported a recognition rate on the Extended Yale-B face database of 89.7\%  that approaches the recognition rate of 90.2\% obtained from the original HR images. Juefei-Xu \etal~\cite{2016:Juefeixu} employed an attention model that shifts the network's attention during training by blurring the images by different amounts to handle gender prediction. An improved version of \cite{juefeixu_2017} was proposed the following year which was a generative approach for occluded image recovery. The improved version boosted performance on gender classification.

\paragraph{Simultaneous super-resolution and face recognition}
\label{Sec:Simultaneous_SRFR}

%-AMYWorks demonstrated the supreme of super-resolution and learning good representation and metrics simultaneously can be found in  \cite{ huang2011super, 2012:Zou,2015:Jian, shekhar2017synthesis}.% hennings2008simultaneous,jia2005multi,arandjelovic2007manifold,}
The main contribution of this group of work is that the recognition oriented constraint is embedded into the super-resolution framework so that the face images reconstructed are well separated according to their identities in the feature space. Compared to approaches mentioned in the previous section, fusing super-resolution and recognition in the same process forces the model to learn more inter-class variations in addition to intra-class similarities.
Zou~\etal~\cite{2012:Zou} stated that traditional example-based or map-based super-resolution approaches tended to hallucinate the LR face images by minimizing the reconstruction error in the LR space using objective function in (\ref{Eq:Zou_1}), where ${\bf I}_h$ and ${\bf I}_l$ are HR and LR images respectively and ${\bf D}$ is the dictionary:   
\begin{equation}
\left \| {\bf D}{\bf I}_h-{\bf I}_l \right \|^{2} \rightarrow \min.
\label{Eq:Zou_1}
\end{equation}
When applied to the LQFR task, these methods tend to generate HR faces that look like the LR faces, but the similarity measured in the LR face space cannot hold enough information to reflect the similarity (intra-class) or saliency (inter-class) of the face in HR face space;  thus, serious artifacts were introduced. Zou~\etal addressed the problem by transferring the reconstruction task from the LR face space to the HR face space (\ref{Eq:Zou_2}) with a regressor ${\bf R}$: 
\begin{equation}
\left \| {\bf R}{\bf I}_l-{\bf I}_h \right \|^{2}.
\label{Eq:Zou_2}
\end{equation}
Moreover, a label constraint is also applied during the optimization. To handle overfitting, a clustering-based algorithm is proposed. Later, Jian~\etal~\cite{2015:Jian} discovered and proved that singular values are effective to represent face images. The singular values of a face image at different resolutions are approximately proportional to each other with the magnification factor and the largest singular value of the HR face images and the original LR image can be utilized to normalize the global feature to form scale invariant feature vectors:
\begin{equation}
s^{'}_{h}=\frac{s_{h}}{w^{1}_{h}}\ \ \text{and} \ \ s^{'}_{l}=\frac{s_{l}}{w^{1}_{l}} 
\label{Eq:Jian}
\end{equation}
where~$w^{1}_{l}$ is the largest singular component of the singular value vector $s^{'}_{h}$. With the scale invariant feature, the proposed method incorporated the class information by learning two mapping matrices from HR-LR images pairs in the same class. The HR faces can be then reconstructed from the two mapping matrices learned. 
Shekhar~\etal~\cite{shekhar2017synthesis} presented a generative approach for LQFR. An image relighting method was used for data augmentation to increase robustness to illumination changes. Following this, class-specific LR dictionaries are learned trough K-SVD and sparse constraints with the reconstruction error being minimized in the LR domain. In the testing phase, the LR probes are projected onto the span of the atoms in the learned dictionary using the orthogonal projector and a residual vector is calculated for each class for the identity classification. The author extended the generic dictionary learning into non-linear space by introducing kernel functions. 
\paragraph*{Deep learning}
Prasad~\etal~\cite{prasad2018genlr} and Zhang~\etal~\cite{zhang2018super} designed deep network architectures and customized the optimization algorithm to tackle the cross-resolution recognition task. Prasad~\etal, in his work, explored different kinds of constraints at different stages of the architecture systematically. Based on the result, they proposed an inter-intra classification loss for the mid-level features combined with super-resolution loss at the low-level feature in the training procedure. Zhang~\etal incorporate an identity loss to narrow the identity difference between a hallucinated face and its corresponding high-resolution face within the hypersphere identity metric space. They illustrated the challenge in the optimization process using this novel loss as well as giving a domain-integrated training strategy as a solution.

\subsection{Discussion}
As one of the intuitive ways to enhance image quality, super-resolution techniques have generated impressive results. As reviewed in Section \ref{Sec:SR}, sparse representation-based methods and interpolation-based methods are two general approaches to solve this task. Constraints were designed on the reconstruction of the face image to get as much information and the optimization usually takes place in in HR space. Dictionary learning and deep neural networks are two of the most popular techniques. The direct output of these methods are super-resolved face images which can be used for both human visual evaluation and as inputs to an automatic face recognition system. The approaches based on development of a unified space reviewed in Section \ref{Sec:UnifiedSpace} may overlap with some of the super-resolution methods. However, they are easier to design for recognition-robust feature extractions since optimization can happen in both  LR and HR domains simultaneously during the learning of the embedding function. However, most of  the  existing  visual quality driven SR methods are optimized to maximize the PSNR or SSIM metrics, which do not measure the recognition performance directly. A potential effort could be made to derive improved methods and optimizing strategies to generating both high visual quality output and recognition-robust features.

\section{LQ robust feature}
\label{Sec:LowResolutionRobustFeature}
\subsection{Background}
Resolution (or quality)-robust features have been studied in the past decade. Wang~\etal introduced earlier works from 2008 in~\cite{wang_2014}. As mentioned in Sec~\ref{Sec:LQFR}, many factors need to be taken into consideration when designing these features. Misalignment, pose and lighting variation are the most important factors that degrade general face recognition performance. In addition, the robust features need to be resolution or quality invariant when conducting cross-resolution (or quality) face matching. The approaches to obtain these features are generally not learning-based, and most of the features are handcrafted and texture or color-based, globally or locally. 
\subsection{Methods}
Because gait features are invariant to the previously mentioned factors and can be collected at a distance without the collaboration of a subject, they have been used to provide aid to various tasks. One work employing gait features, Ben \etal~\cite{ben2012gait}, proposed a feature fusion learning method to couple the LR face feature and gait feature by mapping them into a kernel-based manifold to minimize the distance between the two types of features extracted from the same individual. They achieved better performance on a constructed database containing data from the ORL database \cite{jain2011handbook} and the CASIA(B) gait database \cite{yu2006framework}, compared to the previous methods like CLPM \cite{2010:Li} and Hunag's \cite{huang2011super} RBF-based method.
%[][Amy, the ORL data set does not include gait - how do they use it here?]
%Cui~ \etal \cite{2013:Cui} proposed a method for face recognition in the wild and tried to solve misalignment by dividing the face into spatial blocks and extract sparse features from the position-free patches sampled within each block. A sum-pooling is applied to the extracted sparse representation. They formed a new distance metric learning method to refine the discriminative power of the model. The model is tested with LFW and YTF data set and achieved accuracy with 79.48\% compared with normal LBP with 65.7\%.

LBP features are widely applied in face recognition and have proven to be highly discriminative, with their key advantages, namely, their invariance to monotonic gray-level changes and computational efficiency. Herrmann \cite{2013:Herrmann} addressed the LQFR in the video by suggesting several modifications based on local binary pattern (LBP) features for local matching. They avoided the sparse LBP-histograms in the small local regions on low-resolutions faces by using different scales and temporal fusion with head poses. They achieve recognition rate with 84\% on the HONDA data set and 78\% on VidTIMIT data set with face images downsampled to 8 \rx 8 in pixels. Kim \etal  \cite{2014:Kim} extracted the LBP features adaptively by changing the radius of the circular neighborhood based on the images' sharpness, as measured by kurtosis. It achieved up to 7\%  recognition performance gain on the CMU-PIE data set and 5\% on BU-3DFE data set compared to using the original LBP features.

To overcome the variation in lighting condition, Zou \etal \cite{2013:Zou} proposed a GLF feature which is represented in an illumination insensitive feature space. Using the GLF feature for tracking, faces from 20 \rx 20 to 200 \rx 200 could be successfully tracked on their videos captured outdoors. They also evaluated their tracker on Dudek previously used by Ross \etal \cite{ross2008incremental} and the Boston University head tracking database \cite{la1999fast}.%[WHICH ND data set?]

In 2013, Khan~\etal~\cite{2013:Khan} and Mostafa~\etal~\cite{2013:Mostafa} conducted a study on LR facial expression recognition and LR face recognition in thermal images. Khan~\etal combined a pyramidal approach to extract LBP features at different resolution only from the salient region of a face. 

Mostafa~\etal chose Haar-like features and the AdaBoost algorithm as the baseline for thermal face recognition and improved earlier work~\cite{martinez2010facial}, producing a texture detector probability which was later combined with a shape prior model. They also introduced a normal distribution method to make the shape rotation variant. They reported performance on the ND-data-X1 thermal face image data set using their keypoint detector and a series of texture feature extraction methods. Above 85\% recognition rate was achieved when the thermal face images are of size 16x16 pixels.

In 2015, Hernandez \etal  \cite{2015:Hernandez} argued that using the dissimilarity representation was better than utilizing traditional feature representation and they yield an experimental conclusion that when training images are down-scaled and then up-scaled while the test images are up-scaled, the multi-dimensional matching performance was the best approach to use.
A system based on improving local phase equalization~\cite{2017:Xiao} was proposed by Xiao~\etal They improved the original LPQ in several aspects: First, they extend LPQ to ``LPQ plus" by directly quantizing the Fourier transformation of the blurry image densely. To increase the discriminative power of LPQ, they embedded LPQ into the Fisher vector extraction procedure. Finally, square root and L2 normalization steps were added before using an SVM for classification. Face recognition rates were reported on the Yale database with Gaussian blurring at different levels. With the largest blurring kernel, the proposed feature representation outperformed LBP by nearly 50\% and LPQ by 30\%.
In contrast to most existing works aiming at extracting multi-scale descriptors from the original face images, Boulkenafet~\etal~\cite{2016:Boulkenafet} derived a new multi-scale space through three multiscale filtering steps, including Gaussian scale space, the difference of Gaussian scale space and multiscale Retinex. This new feature space was used to represent different qualities of face images for a face anti-spoofing task. LBP features are also employed in this work.
Peng~\etal~\cite{2016:Peng} enhanced the feature by introducing matching-score-based registration and extended training set augmentation.
El Meslouhi~\etal~\cite{2017:El} used Gabor filter and HOG based feature for feature fusion and the resulting feature was sent to a 1D Hidden Markov Model. Compared with other HMM model that each stage are defined with a region on the face image, this model estimated the parameters during the training step without knowing any prior information of interested regions.

\subsection{Discussion}
From the methods reviewed in Section \ref{Sec:LowResolutionRobustFeature}, the features designed for the LQFR task mostly are texture-based features such as LBP, LPQ or HOG. Multi-scale processing, feature fusion or matching score post processing are employed to enhance the features to be quality (resolution) invariant. They are fast and training free compared to those learning-based methods but it has its limitation when there is less texture information captured in LR face images. In addition, the hand-crafted features are very sensitive to pose, illumination occlusion and expression change.

\section{Unified space}
\label{Sec:UnifiedSpace}
\subsection{Background}
In contrast to designing super-resolution algorithm or LR robust features, there is a group of methods that focuses on learning a common space from both LR probe images and HR gallery images. In the testing phase, LR probes and HR gallery images are projected to the learned space for similarity measurement. These spaces could be linear or non-linear depending on the model definition.
\subsection{Methods}
Similar to the work of Li~\etal~\cite{2010:Li}, Biswas~\etal~\cite{2012:Biswas} introduced a multidimensional scaling method to simultaneously embed the LR probes and HR gallery images into the common space so that the distance between the probe and its counterpart HR approximates the distance between the corresponding high-resolution images.  They achieved recognition rate at 87\% when HR probe images were downsampled to 15 \rx 12 pixels and 53\% when downsampled to 8\rx 6 pixels. They also achieved a prominent result on the SCface data set, achieving 76\% recognition rate using a LBP descriptor. Biswas~\etal~\cite{2013:Biswas} also extended their method to handle uncontrolled poses and illumination conditions. They introduced a tensor analysis to handle rough facial landmark localization in the low-resolution uncontrolled probe images for computing the features. Wang~\etal~\cite{2015:WangYang} treated the LR and HR images as two different groups of variables, and used CCA to determine the transform  between them. They then project the LR and HR images to the common linear space which effectively solved the dimensional mismatching problem. Almaadeed~\etal~\cite{2016:Almaadeed} represented LR faces and HR faces as patches and proposed a dictionary learning method so that the LR face can be represented by a set of HR visual words from the learned dictionary. A random pooling is applied on both HR and LR patches for selecting a subset of visual words and K-LDA is applied for classification. Wei~\etal~\cite{2017:Wei} developed a mapping between HR and LR faces using linear coupled dictionary learning with sparse constraint to bridge the discrepancy between LR and HR image domains. Heinsohn~\etal~\cite{Heinsohn2019:IMAVIS} proposed a method, called blur-ASR or bASR, which was designed to recognize faces using dictionaries with different levels of blurriness and have been proven to be more robust with respect to low-quality images.

Some methods not only focus on intra-class similarity but also designed models that emphasize the inter-class variation for more discriminative features. Shekhar~\etal~\cite{2011:Shekhar} present a generative-based approach to low-resolution FR based on learning class specific dictionaries, considering pose, illumination variations and expression as parameters. Moutafis~\etal~\cite{2014:Moutafis} jointly trained two semi-coupled bases using LR images with class-discriminated correspondence that enhance the class-separation. Lu~\etal~\cite{lu2016very} proposed a similar semi-coupled dictionary method using locality-constrained representation instead of sparse representation and applied sparse representation based classifier to predict the face labels. Compared with~\cite{2010:Li}, Jiang~\etal and Shi~\etal~\cite{2015:Jiang, 2015:Shi} improved the coupled mapping by simultaneously learning from neighbor information and local geometric structure within the training data to minimize the intra-manifold distance and maximize the inter-manifold distance so that more discriminant information is preserved for better classification performance. Zhang~\etal~\cite{2015:Zhang} similarly proposed a coupled marginal discriminant mapping method by employing a inter-class similarity matrix and within-class similarity matrix. Similarly, Wang~\etal~\cite{2016:WangHu} addressed on pose and illumination variations for probe and gallery face images and propose a kernel-based discriminant analysis to match LR non-frontal probe images with HR frontal gallery images. Yang~\etal~\cite{2017:Yang} employed multidimensional scaling (MDS) to handle HR and LR dimensional mismatching. Compared with previous methods using MDS, such as~\cite{2012:Biswas}, they enhanced the model's discriminative power by introducing an inter-class-constraint to enlarge the distances of different subjects in the subspace. Yang~\etal~\cite{yang2018discriminative} also proposed a more discriminative MDS method to learn a mapping matrix, which projects the HR images and LR images to a common subspace. It utilizes the information within HR-LR,HR-HR, and LR-LR pairs and set an inter-class constraint is employed to enlarge the distances of different subjects in the subspace to ensure discriminability. Mudunuri~\etal~\cite{mudunuri2016low} introduced a new MDS method but improved it to be more computationally efficient by using a reference-based strategy. Instead of matching an incoming probe image with all gallery images, it only matches probe face images to selected reference images in the gallery. 
Mudunuri~\etal~\cite{2017:Mudunuri} constructed two orthogonal dictionaries in LR and HR domain and aligned them using bipartite graph matching. Compared to other methods, this method does not need one-to-one paired HR and LR pairs. Similar to~\cite{2017:Mudunuri}, Xing~\etal~\cite{2016:Xing} proposed a bipartite graph based manifold learning approach to build the unified space for HR and LR face images. This method contained construct the graph on two sample set namely HR and LR and used coupled manifold discriminant analysis to embed the HR and LR face images into the same unified space for matching. 

Haghighat~\etal~\cite{haghighat2017low} proposed a Discriminant Correlation Analysis (DCA) approach to highlight the differences between classes, as an improvement over Canonical Correlation Analysis.
Gao~\cite{gao2018low} mentioned that most previous works ignored the occlusions in the LR probe. They used double low-rank representation to reveal the global structures of images in the gallery and probe sets which might have occlusion. The nuclear norm is used to characterize the reconstruction error. A sparse representation based classiﬁer is also used for classification.
\paragraph*{Deep learning}
The method of Dan~\etal~\cite{2016:Danzeng}, Li~\etal~\cite{li_2018}, and Lu~\etal~\cite{lu2018deep} is bases on deep learning. Dan~\etal~\cite{2016:Danzeng} designed a model using deep architecture and mixed LR images with HR images for training to learn a highly non-linear resolution invariant space. Li~\etal~\cite{li_2018} improved \cite{2016:Danzeng} by introducing a centerless regularization to further push the intra-class samples closer in the learned representation space. Similar to Li~\etal~\cite{li_2018} strategies, Lu~\etal~\cite{lu2018deep} introduced an extra branch network on top of a trunk network and proposed a new coupled mapping loss performing supervision in the feature domain in the training process. They achieved 73.3\% of rank one rate on the largest standoff set of SCface data set.

\subsection{Discussion}
The unified space learning methods for LR and HR face images share some strategies with SR methods  employing CCA, sparse representation learning and bipartite graph based manifold learning. However, they tend to focus on generating features for cross-resolution matching. Since performance is measured by recognition rate, most of the methods are designed to obtain more discriminative and class-separable features when the LR and HR images are mapped into the common space.

\section{Deblurring}
\label{Sec:Deblurring}
\subsection{Background}
Blurriness can exist in many LQ faces captured in uncontrolled scenes, and can be considered to  degrade the recognition performance as if the face image is very small in size. Most existing image deblurring methods fall into one of these categories:
\begin{itemize}

\item {\em Blind image deblurring (BID) and Non-blind image deblurring (NBID)} --
BID techniques reconstruct the clear version of the image without knowing the type of blurriness. In this case, there is no prior knowledge available except the images. NBID techniques employ a known blurring model.

\item {\em Global deblurring and local deblurring} -- global blurriness can happen when the cameras are moving quickly or the exposure time is long. Local blurriness only influences a portion of an image. Compared with local deblurring, global deburring only needs to estimate the blur model parameters (such as PSF estimation) to restore the image. For local blurriness, it is more challenging since the image may be influenced by several blurring kernels which need to be modeled (including their region of effect), and their parameters estimated, from the image for restoration to occur.

\item  {\em Single image deblurring and multi-image deblurring} -- single image deblurring uses only one image as an input while multi-image deblurring uses multiple images.  Multiple images can provide additional information to the deblurring approach such as variations in quality and other imaging conditions.
%The single image deblurring task is more common in a real setting, as supportive information such as a certain frame with better quality or exposure information, which is hard to obtain for non-professional users. 
\end{itemize}
\subsection{Methods}
Blind image restoration has been widely studied over the past few years; methods have focused on estimation of blurring kernels, and have been shown to be effective. Works including~\cite{2011:Nishiyama, 2014:Li, 2015:Li,2016:Tian, 2016:Kumar} use PSFs to estimate and restore well-modeled blur contamination.
Nishiyama~\etal~\cite{2011:Nishiyama} defined a set of PSFs to learn prior information in the training process using a set of blurred face images. They constructed a frequency-magnitude-based feature space which is sensitive to appearance variation of different blurs to learn the representation for PSF inference and apply it to new images. They also show that LPQ can be combined with the proposed method to yield a better result. Li~\etal~\cite{2014:Li} also combined their method with subspace-based point spread function (PSF) estimation method to handle cases of unknown blur degree, and adopted multidimensional scaling to learn a transformation with HR face images and their blurry counterparts for face matching. Li~\etal's method~\cite{2015:Li} adaptively determined a sample frequency point for a specific blur PSF that managed the tradeoff between noise sensitivity and classification performance.
Tian~\etal\cite{2016:Tian} proposed a method that defined an estimated PSF with a linear combination of a set of pre-defined orthogonal PSFs and an intrinsic sharp image (EI) that consisted of a combination of a set of pre-defined orthogonal face images. The coefficients of the PSF and EI are learned jointly by minimizing a reconstruction error in the HR face image space, and a list of candidate PSFs is generated. Finally, they used BIQA-based approach to select the best image from the results processed by the filters on the candidacy list. Kumar \etal~\cite{2016:Kumar} discovered that based on the PSF shape, the homogeneity and the smoothness of the blurred image in the motion direction are greater than in other directions. This could be used to restore the image for identification.
Lai~\etal~\cite{2016:Lai} presented the first comprehensive perceptual study and analysis of single image blind deblurring using real-world blurred images and contributed a data set of real blurred images and another data set of synthetically blurred images.
Zhang \etal~\cite{2011:Haichao} proposed a joint blind image restoration and recognition method based on a sparse representation prior when the real degradation model is unknown. They combined restoration and recognition in a unified framework by seeking sparse representation over the training faces via L1 norm minimization. Heinsohn \etal~\cite{2015:Heinsohn} tried to solve the problem by using an adaptive sparse representation of random patches from the face images combined with dictionary learning
Vageeswaran \etal~\cite{2013:Vageeswaran} defined a bi-convex set for all images blurred and illumination altered from the same image and solve the task by jointly modeling blur and illumination.
Mitra \etal~\cite{2014:Mitra} proposed a Bank-of-Classifiers approach for directly recognizing motion blurred face images. 
Some methods are based on priors such as facial structural or facial key points. Pan \etal~\cite{2014:Pan} proposed a maximum a posteriori (MAP) deblurring algorithm based on kernel estimation from the example images' structures, which is robust to the size of the data set. Huang \etal~\cite{2015:Huang} introduced a computationally efficient approach by integrating classical L0 deblurring approach with face landmark detection. The detected contours are used as salient edges to guide the blind image deconvolution.
Flusser \cite{2016:Flusser} proved that the primordial image is invariant to Gaussian blur which exists in most LR face images. They used the central moments and scale-normalized moments to handle rotation and scaling variance which also applied to the uncontrolled condition of face images captured under surveillance cameras. They reported that the new feature outperformed Zhang's distance and local phase equalization.
\paragraph*{Deep learning}
Most recent deep learning-based face deblurring approaches perform end-to-end deblurring without specific modeling for blurriness kernel estimation. Dodge \etal~\cite{2016:Dodge} provide an evaluation of four state-of-the-art deep neural network models for image classification under five quality distortions including blur, noise, contrast, JPEG, and JPEG2000 compression, showing that the VGG-16 network exhibits the best performance in terms of classification accuracy and resilience for all types and levels of distortions as compared to the other networks. Pherson~\etal~\cite{2016:McPherson} first presented obfuscation techniques to remove sensitive information from images including mosaicing blurring and, a proposed system called P3~\cite{ra2013p3}. The proposed supervised learning is performed on the obfuscated images to create an artificially obfuscated-image recognition model. Chrysos \etal~\cite{2017:Chrysos} utilized a Resnet-based non-max-pooling deep architecture to perform the deblurring and employ a face alignment technique to pre-process each face in a weak supervision fashion. Jin \etal~\cite{jin2018learning} improve blind deblurring by introducing a scheme called re-sampling that generate larger reception field in the early convolutional layer and combine it with instance normalization which is proved outperform batch normalization at the deblurring task. In Shen \etal~\cite{shen2018deep}, global semantic priors of the faces and local structure losses are exploited in order to restore blurred face images with a multi-scale deep CNN.

\subsection{Discussion}
Most methods reviewed in this section introduce prior information such as the point spread function (PSF) and try to predict an estimation of the blurring kernel for converting the image back to clear image. There also are novel works on predicting random blurring kernel without prior information.

%\subsubsection{Discussion}
%These deep learning methods provide a more elegant way to unify each unit in the face reconstruction or face recognition pipeline and optimize them simultaneously for a better visual quality result or feature representation and recognition rate. However, not enough studies have been conducted on the intrinsic representation capacity for LR face images and thus there is no clear theoretical guidance for designing new architectures to specifically fit this task.

\section{Data Sets and Evaluations}
\label{Sec:data setEvaluations}

In this section, we carefully review the data sets, experimental settings, evaluation protocols, and performance of some of the methods discussed in Section \ref{Sec:LQFR}. We address the research effort on the LQFR problem in two aspects: general LQFR techniques which studied artificially generated LR face images, and LQFR research that employed LR face images captured in the wild, motivated by real life scenarios such as surveillance. 
%LR face recognition is a task to solve matching LR probe images to HR gallery images to establish identities. A general way to do this is to perform 1 to N matching, which matches the template (feature vector) extracted from the probe image to the templates of each of the images in the gallery and rank the distance of each image pairs to get the best candidate. 
There are many data sets being employed to LQFR research topic, and each may include its own protocols for performance evaluation. To give a better understanding of the performance of the methods we reviewed in the previous section, we choose several data sets which are evaluated in many works and compare the results reported.
%In order to illustrate the performance difference between constrained synthetic LR faces and LR faces captured under an unconstrained environment such as surveillance cameras.
A summary of the data sets used in previous works is shown in Table \ref{Tab:DataSum}.

\subsection{Data Sets Collected Under Constrained Conditions}
\label{Sec:Constraineddata sets}

Most of the works that address LQFR generate LR versions of HR face images by re-sizing (downsampling) the HR face to a smaller scale and sometimes adding a defined blurriness amount (Gaussian or motion blur). The original face images from these data sets are all collected under constrained environments by HD cameras and are studied after delicate facial alignment. We choose three data sets used in previous works in order to give a fair comparison with different methods. The gallery image size and probe image size were defined as shown in Table \ref{Tab:PORF_CONST}, some of the work might define a set of experiment settings in order to research performance on resolution change. We selected the most representative settings to report below. Methods proposed by Lee ~\etal~\cite{kim_2014} and Yang~\etal~\cite{2017:Yang} all achieved recognition rates above 90\% on the FRGC data set and FERET data sets when the probe image size was as low as 8 \rx 8 pixels.

When pose and illumination variations were introduced into the matching problem, matching performance should be expected to degrade. Some specially designed pose-robust methods such as the method of Biswas~\etal~\cite{2013:Biswas} and Wang~\etal~\cite{2016:WangHu} explored the impact of pose and illumination change on recognition performance on the Multi-PIE data set. While the techniques developed were more robust than prior methods, recognition rate still dropped (for example, the method of Biswas \etal~\cite{2013:Biswas} dropped more than 20\%, and Wang \etal's method~\cite{2016:WangHu} by 15\% under Multi-PIE illumination condition 10 with significant nonfrontal face pose. 
\begin{table*}[]
\centering
\caption{Performance evaluation on standard constrained face data set}
\label{Tab:PORF_CONST}
\scalebox{0.8}{
\begin{tabular}{lcccc}
\toprule
                               & Method and                   & Gallery resolution  & Probe resolution                 & Recognition rate$^*$ \\ 
data set                        & reference                    & [pixels]            & [pixels]                         & [\%]             \\ \midrule
\multirow{7}{*}{FRGC}          & \cite{2012:Zou}              & 56 \rx 48           & 7 \rx 6                          & 77.0             \\ \cline{2-5} 
                               & \cite{2012:Gopalan}          & 64 \rx 64           & Random real-world image blur     & 84.2             \\ \cline{2-5} 
                               & \cite{kim_2014}           & --                  & --                               & 95.2             \\ \cline{2-5} 
                               & \cite{2015:Li}               & 130 \rx 150         & Random real-world image blur     & 12.2             \\ \cline{2-5} 
                               & \cite{2017:Farrugia}         & 8$^{**}$            & 40$^{**}$                        & 60.3             \\ \cline{2-5} 
                               & \cite{haghighat2017low}      & 128 \rx 128         & 16 \rx 16                        & 100.0            \\ \cline{2-5} 
                               & \cite{shekhar2017synthesis}  & 48 \rx 40           & 7 \rx 6                          & 75.0             \\ \midrule
\multirow{11}{*}{FERET}        & \cite{uiboupin2016facial}    & 60 \rx 60           & 15 \rx 15                        & 21.6             \\ \cline{2-5} 
                               %& \cite{2010:Li}               & 72 \rx 72           & 12 \rx 12                        & 90.1             \\ \cline{2-5} 
                               %& \cite{2011:Nishiyama}        & 128 \rx 128         & 64 \rx 64 Random artificial blur & 82.9             \\ \cline{2-5} 
                               & \cite{2012:Gopalan}          & 64 \rx 64           & Random artificial blur           & 97.1             \\ \cline{2-5} 
                               & \cite{2013:Vageeswaran}      & --                  & Random Gaussian blur             & 93.0             \\ \cline{2-5} 
                               & \cite{2014:Li}               & 128 \rx 128         & 64 \rx 64 Gaussian blur          & 88.6             \\ \cline{2-5} 
                               & \cite{2015:Shi}              & 32 \rx 32           & 8 \rx 8                          & 81.0             \\ \cline{2-5} 
                               & \cite{2015:Zhang}            & 80 \rx 80           & 10 \rx 10                        & 88.5             \\ \cline{2-5} 
                               & \cite{2015:Li}               & 130 \rx 150         & Random artificial blur           & 79.6             \\ \cline{2-5} 
                               & \cite{2016:Rasti}            & 60 \rx 60           & 15 \rx 15                        & 21.6             \\ \cline{2-5} 
                               & \cite{2017:Yang}             & 32 \rx 32           & 8 \rx 8                          & 93.6             \\ \midrule
\multirow{9}{*}{CMU-PIE}       & \cite{2012:Zou}              & --                  & --                               & 74.0             \\ \cline{2-5} 
                               & \cite{2013:Vageeswaran}      & --                  & Random Gaussian blur+illumination change & 81.4     \\ \cline{2-5} 
                               & \cite{2014:Li}               & 128 \rx 128         & 64 \rx 64 Gaussian blur          & 87.4             \\ \cline{2-5} 
                               & \cite{2014:Kim}              & 48$^{**}$           & Random artificial blur           & 66.9             \\ \cline{2-5} 
                               & \cite{shekhar2017synthesis}  & 48 \rx 40           & 7 \rx 6                          & 78.0             \\ \cline{2-5} 
                               & \cite{2015:Jiang}            & 32 \rx 28           & 8 \rx 7                          & 98.2             \\ \cline{2-5} 
                               & \cite{2016:Xing}             & 32 \rx 32           & 12 \rx 12                        & 93.7             \\ \cline{2-5} 
                               & \cite{2016:Tian}             & --                  & Random airtificial blur          & 95.0             \\ \cline{2-5} 
                               & \cite{2017:Wei}              & 16 \rx 16           & 8 \rx 8                          & 80.0             \\ \midrule
\multirow{8}{*}{CMU Multi-PIE} & \cite{2011:Haichao}          & 80 \rx 60           & Random airtificial blur          & 61.3             \\ \cline{2-5} 
                               & \cite{2012:Biswas}           & 48 \rx 40           & 8 \rx 6                          & 53.0             \\ \cline{2-5} 
                               & \cite{2013:Biswas}           & 65 \rx 55           & 20 \rx 18                        & 89.0             \\ \cline{2-5} 
                               & \cite{2014:Moutafis}         & 24 \rx 24           & 12 \rx 12                        & 80.4             \\ \cline{2-5} 
                               & \cite{2015:Shi}              & 32 \rx 32           & 8 \rx 8                          & 95.7             \\ \cline{2-5} 
                               & \cite{2016:WangHu}           & 60 \rx 60           & 20 \rx 20                        & 89.0             \\ \cline{2-5} 
                               & \cite{2017:Mudunuri}         & 36 \rx 30           & 18 \rx 15                        & 96.3             \\ \cline{2-5} 
                               & \cite{2017:Yang}             & 32 \rx 32           & 8 \rx 8                          & 95.8             \\ \bottomrule
\end{tabular}
}

\small{$^*$ Not comparable because evaluation protocols are different.}

\small{$^{**}$ Number of pixels between eyes.}

\end{table*}

\subsection{Data Sets Collected Under Unconstrained Conditions}
\label{Sec:Unconstraineddata sets}

The unconstrained face data sets are captured without the subjects' cooperation, yielding random poses (sometimes extreme poses), varying resolutions and different subjective quality levels.
In Table~\ref{Tab:PORF_UNCONST}, we summarize face data sets available for research purposes, which were collected either through surveillance cameras or other digital devices in an unconstrained environment. %At the first glance of performance and compare with that evaluation performance on data set under constrained environment,
As expected, the performance of all the methods evaluated on these unconstrained data sets was much lower than performance of methods that employed images captured in constrained conditions. Cheng~\etal~\cite{2018:Cheng}\cite{2018:Cheng2} present two large LR face data sets assembled from existing public face data sets. They conducted baseline experiments on these data sets and (similar to Li~\etal~\cite{li_2018}) also compared the recognition performance gap between synthetic LR face image and native unconstrained LR face images. They identified face image quality control and good face alignment as key challenges to performance.
%Not like artificially generated LR faces, which although are degraded but still contain a reasonable face region and were aligned when it is up-scaling back via super-resolution or other techniques. However,
%Face images captured by surveillance cameras and other digital devices with large standoffs is both hard for face detection and face alignment.
Recent works such as~\cite{bulat2017super} and~\cite{chen2017fsrnet} employed deep learning methods for LQFR.
which optimize the whole system in an elegant end-to-end fashion.
They incorporated facial contour and landmark estimation for LR faces while training which was intended to improve the quality of face alignment. They also showed a competitive visual qualitative result for the super-resolution output using LR faces from a large unconstrained data set collected from the web. Precise recognition rates are not reported.
%, their present a competitive result on enhance LQ face images in unconstrained environment.

Another challenge for unconstrained LQFR is the training process. Since it is hard to collect and ground truth surveillance video, algorithms trained on artificially generated LR and HR pairs cannot capture the real world LR face distribution. This may result in degraded performance on some data sets.
%such as SCface data set.
In addition, with LQ faces detected by state-of-the-art face detection algorithms in surveillance-like quality video frames, HR images of the same subjects may be difficult or impossible to obtain.
%This create the challenge to most super-resolution based LQFR approaches when they try to explore real-life data.

%%%%%%%%%%%%%%%%%%%%%%%%%%%%%%%%%%%%%%%%%%%%%%%%%
\begin{table*}[]
\centering
\caption{Performance evaluation on unconstrained face data set}
\label{Tab:PORF_UNCONST}

\begin{tabularx}{1\textwidth}{lcccc}
\hline
data set Name            & Method                  & Gallery resolution & Probe resolution  & Recognition rate$^*$   \\ 
data set                 & reference               & [pixels]           & [pixels]          & [\%]             \\ \midrule
\multirow{8}{*}{SCface} & \cite{2012:Biswas}      & --                 & --                & 76.0          \\ \cline{2-5} 
                        & \cite{2014:Moutafis}    & 30 \rx 24          & 15 \rx 12         & 52.7          \\ \cline{2-5} 
                        & \cite{2015:Shi}         & 48 \rx 48          & 16 \rx 16         & 43.2          \\ \cline{2-5} 
                        & \cite{2016:Danzeng}     & --                 & --                & 74.0          \\ \cline{2-5} 
                        & \cite{2017:Mudunuri}    & --                 & --                & 73.3          \\ \cline{2-5} 
                        & \cite{2017:Yang}        & 48 \rx 48          & 16 \rx 16         & 81.5          \\ \cline{2-5} 
                        & \cite{haghighat2017low} & 128 \rx 128        & 16 \rx 16         & 12.2          \\ \cline{2-5} 
                        & \cite{2015:Jian}        & 72 \rx 64          & 18 \rx 16         & 52.6          \\ \cline{2-5}
                        & \cite{2012:Zou}         & 64 \rx 56          & 16 \rx 14         & 22.0          \\ \cline{2-5}
                        \midrule
UCCSface                & \cite{2016:Wang}        & 80 \rx 80          & 16 \rx 16         & 59.0          \\ \cline{2-5}
                        \midrule
LFW                     
                        
                        & \cite{2015:Jian}        & 72 \rx 64          & 12 \rx 14         & 66.19          \\ \cline{2-5}

                        & \cite{zhang2018super}        & 112 \rx 96          & 12 \rx 14         & 98.25          \\ \cline{2-5}    
                        & \cite{prasad2018genlr}        & 224 \rx 224          & 20 \rx 20         & 90     \\ \cline{2-5}

                        \midrule
YTF                     
                        
                        & \cite{zhang2018super}        & 112 \rx 96          & 12 \rx 14         & 93     \\\cline{2-5}
                        \midrule
CFP                     & \cite{prasad2018genlr}        & 80 \rx 80          & 20 \rx 20         & 77.28          \\ \cline{2-5}

QMUL-TinyFace           & \cite{2018:Cheng2}        & --          & 20 \rx 16         & 77.28          \\ \cline{2-5}

\bottomrule
\end{tabularx}

\small{$^*$ Not comparable because evaluation protocols are different.}

\end{table*}
% \usepackage{multirow}
\begin{table*}[ht]
\caption{Dataset Summary}
\label{Tab:DataSum}
%\begin{adjustbox}{width=\textwidth}
\renewcommand{\arraystretch}{2}
\scalebox{0.53}{
\begin{tabular}{cccccccc}
\hline
                               & Name          & Source                               & Quality  & Video/static/3D  & Number of subject            & Number of total images      & Variations                                \\ \hline
\multirow{10}{*}{constrained}  & FRGC          & MC                   & HR       & static/3D        & 222(training)466(validation) & 12,776                      & background                                \\ \cline{2-8} 
                               & CMU-PIE       & MC                   & HR       & static           & 68                           & 41,368                      & pose, illumination, expression            \\ \cline{2-8} 
                               & CMU-Multi-PIE & MC                   & HR       & static           & 337                          & 750,000                     & pose, illumination, expression            \\ \cline{2-8} 
                               & Yale-B        & MC                   & HR       & static           & 10                           & 5850                        & pose, illumination                        \\ \cline{2-8} 
                               & CAS-PEAL-R1   & MC                   & HR       & static           & 1,040                        & 30,900                      & pose, illumination, accessory, background \\ \cline{2-8} 
                               & CUFS          & sketch+MC & HR       & static           & 606                          & 1,212                        & -                                         \\ \cline{2-8} 
                               & AR            & MC                   & HR       & static           & 126                          & 4,000                       & illumination, expression, occlusion       \\ \cline{2-8} 
                               & ORL           & MC                   & HR       & static           & 40                           & 400                         & illumination, accessory                   \\ \cline{2-8} 
                               & Superfaces    & MC                   & LR+HR    & static/3D/videos & 20                           & 40                          & -                                         \\ \cline{2-8} 
                               & headPose       & MC                   &          & static           & 15                           & 2,790                        & pose, accessory                           \\ \hline
\multirow{7}{*}{unconstrained} & PaSC          & MC                    & HR+blur  & static+video     & 293/265                      & 9,376(static) 2,802(video) & environment                               \\ \cline{2-8} 
                               & SCface        & surveillance                         & HR+LR    & static           & 130                          & 4,160                        & visible and infrared spectrum             \\ \cline{2-8} 
                               & EBOLO         & surveillance                         & LR       & video            & 9 and 213 distractors        & 114,966                      & accessory                                 \\ \cline{2-8} 
                               & QMUL-SurvFace & surveillance                         & LR       & static+video            & 15,573                       & 463,507                     & -                                         \\ \cline{2-8} 
                               & QMUL-TinyFace & Web                         & LR       & static            & 5,139                       & 169,403                     & -                                         \\ \cline{2-8} 
                               & UCCSface      & surveillance                         & HR(blur) & static           & 308                          & 6,337                       & -                                         \\ \cline{2-8} 
                               & YTF           & web                                  & HR       & video            & 1,595                        & 3,425(video sequences)      & -                                         \\ \cline{2-8} 
                               & LFW           & web                                  & HR       & static           & 5,749                         & 13,000                      & - 
                               \\ \cline{2-8} 
                               & CFPW           & web                                  & HR       & static           & 500                         & 6,000                      & pose
                               \\ \hline
\end{tabular}
}
%\end{adjustbox}
\small{
Note: MC stands for manually collected using standard devices
}
\end{table*}

\section{Conclusions}
\label{Sec:Conclusions}

Over the last decade, we have witnessed tremendous improvements in face recognition algorithms. Some applications, that might have been considered science fiction in the past, have become reality now. However, it is clear that face recognition, performed by machines or even by humans, is far from perfect when tackling low-quality face images such as faces taken in unconstrained environments \eg face images acquired by long-distance surveillance cameras. In this paper, the attempt was made to establish the state of the art in face recognition in low quality images.

\subsection{Trends and future research}
Wang~\etal~\cite{wang_2014} noted that the learning of discriminative nonlinear mapping for the LQFR task is a promising idea.
%We witnessed elegant modeling highly nonlinear mapping modeling recent years when deep neural networks have been widely employed. These
Deep learning methods provide a more elegant way to unify each unit in the face reconstruction or face recognition pipeline and optimize them simultaneously for a better
%visual quality result or
feature representation and recognition rate. However, not enough studies have been conducted on the intrinsic representation capacity for LR face images and thus there is no clear theoretical guidance for designing new architectures to specifically fit this task. Advances in this modeling task would have significant influence on the research community.
%It would be a great contribution if interpretative model could be involved to explore the discipline.
In addition, real-world data sets employing LQ faces (especially surveillance video-based data sets) are rather limited in size. More real-world data would support more impactful studies in this field.
%And it is still value to notice that Wang \etal \cite{wang_2014} noted for improving super-resolution processing to satisfy both vision and recognition purpose as well as looking for more class discriminative feature learning method in the LR domain.

\subsection{Concluding remarks}
In this paper, we comprehensively summarizes and reviewed the works on the LQFR problem over the last decade. By categorizing related works that addressed the LQFR task, we provided a moderately detailed review of many state-of-the-art representations, followed by a performance evaluation using well-known research data sets. Finally, we reveal future challenges addressing this problem. 

We conclude that finding techniques to improve face recognition in low-quality images is an important contemporary research topic. Among these techniques, deep learning is very robust against some challenges (\eg illumination, some blurriness, some expressions, some poses, etc.) and yet very poor in other cases (\eg very low-resolution, blurriness, compression, etc.). There is a high level of interest in the scientific world in the recognition of faces in low-quality images due to the promising applications (forensics, surveillance, etc.) that have this as their point of departure. It would be interesting to explore deep learning approaches on low-quality images exploring different architectures on ad--hoc training datasets with millions of images.

\section*{Acknowledgment}
This work was supported by College of Engineering at the University of Notre Dame, and in part by the Seed Grant Program of The College of Engineering at the Pontificia Universidad Catolica de Chile, and in part by Fodecyt-Grant 1191131 from Chilean Science Foundation.
% ONLY 1191131: Thanks Amy :)
% I removed  1161314, it was wrong... sorry :)

%\ifCLASSOPTIONcaptionsoff
%  \newpage
%\fi
% \clearpage
%\bibliographystyle{IEEEtranS}
\bibliographystyle{ACM-Reference-Format}
\bibliography{main}

\end{document}